# The Validity of a Machine Learning-Based Video Game in the Objective Screening of Attention Deficit Hyperactivity Disorder in Children Aged 5 to 12 Years

Zeinab Zakani[1], Hadi Moradi[2, 3], Sogand Ghasemzadeh[1], Maryam Riazi[4], and Fatemeh Mortazavi[2]

[1] Department of Psychology, University of Tehran

[2] Department of Robotics and Artificial Intelligence, University of Tehran

[3] Intelligent Systems Research Institute, SKKU

[4] Department of Electrical and Computer Engineering, Boston University

## Author Note

Zeinab Zakani  https://orcid.org/0009-0005-5622-1843

Hadi Moradi  https://orcid.org/0000-0003-4916-9408

Sogand Ghasemzadeh  https://orcid.org/0000-0003-0897-1568

Maryam Riazi  https://orcid.org/0009-0006-6089-3914

Fatemeh Mortazavi  https://orcid.org/0000-0002-7723-584X

This research was conducted with financial support from the Javaneh Program of the Ministry of Science, Research, and Technology of the Islamic Republic of Iran, and the Cognitive Sciences and Technologies Council of the Islamic Republic of Iran.

Correspondence concerning this article should be addressed to Hadi Moradi, Department of Robotics and Artificial Intelligence, University of Tehran, Tehran, Iran. Email: moradih@ut.ac.ir

**Abstract**

***Objective:*** Early identification of ADHD is necessary to provide the opportunity for timely treatment. However, screening the symptoms of ADHD on a large scale is not easy. This study aimed to validate a video game (FishFinder) for the screening of ADHD using objective measurement of the core symptoms of this disorder.

***Method:*** The FishFinder measures attention and impulsivity through in-game performance and evaluates the child's hyperactivity using smartphone motion sensors. This game was tested on 26 children with ADHD and 26 healthy children aged 5 to 12 years. A Support Vector Machine was employed to detect children with ADHD.

***results:*** This system showed 92.3% accuracy, 90% sensitivity, and 93.7% specificity using a combination of in-game and movement features.

***Conclusions:*** The FishFinder demonstrated a strong ability to identify ADHD in children. So, this game can be used as an affordable, accessible, and enjoyable method for the objective screening of ADHD.

*Keywords:* ADHD, Objective assessment, Serious game, Screening, Artificial intelligence

**The Validity of a Machine Learning-Based Video Game in the Objective**

**Screening of Attention Deficit Hyperactivity Disorder in Children Aged**

**5 to 12 Years**

Attention Deficit Hyperactivity Disorder (ADHD) is one of the most common childhood disorders with a prevalence of about 7.2% (Thomas et al., 2015). ADHD has three core symptoms: attention deficit, impulsivity, and hyperactivity (Banaschewski, 2015).

If ADHD is not diagnosed as early as possible and the treatment is not done on time, there would be many negative consequences in various aspects of a person's life (Shaw et al., 2012). For instance, children and adolescents with ADHD experience many problems such as unfavorable performance in school and academic progress (Arnold et al., 2020), problems in interactions with parents and other family members (Lifford et al., 2009), rejection from peers (binti Marsus et al., 2022), low self-esteem (Harpin et al., 2016), and difficulty in regulating emotions (Christiansen et al., 2019; Moukhtarian et al., 2018). Furthermore, adults with ADHD experience significant defects in work performance and higher levels of unemployment (Küpper et al., 2012), family dysfunction such as higher rates of separation and divorce (Wymbs et al., 2021), and problems in parenting. The possibility of accidents and driving disorders (Brunkhorst-Kanaan et al., 2021), risky sexual behavior (Hua et al., 2021), social and legal problems (Retz et al., 2021), and even suicide attempts (Garas & Balazs, 2020) are also reported to be higher in these individuals.

Despite the severe effects that ADHD has on various aspects of a person's life, treatment at an early age can reduce many of these problems and significantly improve the quality of the person's life (binti Marsus et al., 2022; Cortese et al., 2018; Shaw et al., 2012) Therefore it is necessary to diagnose ADHD as soon as possible to facilitate the start of effective treatments on time (Schoenfelder & Kollins, 2016). Consequently, widely available and automatic ADHD screening methods, that can be used in the absence of an expert, can help in early/on time diagnosis and treatment. Such automatic screening systems may alert parents and caregivers for further evaluation of suspected cases.

On the other hand, the high number of subjective diagnoses of ADHD based on reports from parents, teachers, and guardians regarding children's behavior is a serious challenge (Doulou & Drigas, 2022; Hall et al., 2016). It introduces the need for new methods of objective evaluation of ADHD's symptoms.

Among the various objective methods, the continuous performance test (CPT) is one of the most well-known objective neuropsychological tests used to evaluate ADHD symptoms (Hall et al., 2016). This test has simple instructions that make it usable for children. The subject must select the target stimulus when it is presented and withhold the response when the non-target stimulus is presented. These stimuli are mainly presented in visual forms to evaluate visual attention and response control. Nevertheless, there are tests, like integrated visual and auditory continuous performance test (IVA), that also use auditory stimuli to measure auditory attention and response control in addition to visual attention. Different types of CPT tests vary in terms of the target and non-target stimuli used (numbers, letters, shapes, etc.), the ratio of target to non-target stimuli, the number of stimuli presented, the time of stimulus presentation, the interval between stimulus presentation and the next stimulus, total time test execution, and measured parameters (Hall et al., 2016; Parsons et al., 2019).

CPT tests, such as IVA, Conner's CPT (CCPT), and Test of Variables of Attention (TOVA) mainly assess the state of attention, impulsivity, and inhibition using parameters such as commission errors, omission errors, reaction time, and standard deviation of reaction time. Unfortunately, these tests do not evaluate movement activity during the test to check the presence or absence of hyperactivity and its severity is mostly non-existent.

Hall et al. (2015), in their systematic review of the clinical utility of CPT tests, showed that there is increasing evidence in benefits of using objective measurements of motor activity to evaluate the symptoms of ADHD. The study results indicate that combining a CPT test with an objective measure of activity like the QB test shows potential as a clinical tool, highlighting the need for additional follow-up research.

During recent years, valuable efforts have been made to record movement activity simultaneously with CPT tests (Hall et al., 2016; Murillo et al., 2015). The most important of these efforts, which have also been approved by the FDA, are the quantified behavior test (QbTest) and the McLean motion and attention test (MMAT). In these tests, an infrared camera is used, during a CPT test, to quantify head movement features such as active time, traveled distance, movement area, and micro movements (Emser et al., 2018; Hult et al., 2018).

A weakness of these methods is in the use of infrared cameras which may not be widely available. Furthermore, in the current setups, a headband is used to track subjects which is not a comfortable method and may affect subjects' performance.

More recently, a new method has been proposed to evaluate motor activity simultaneously with the CPT test using webcams. This eliminates the weakness of the need for an infrared camera. Tracking the movements of ADHD patients via a webcam, in addition to the information that is recorded using most CPT paradigms (i.e., errors and reaction time), provides useful information about movement activities to specialists, which significantly contributes to the process of evaluation, diagnosis, and treatment monitoring (Ulberstad et al., 2020).

Another innovative method to assess motor activity during a CPT test is the use of movement sensors placed in a virtual reality headset. Aula test uses these sensors and subjects take a CPT test in a simulated classroom environment (Areces et al., 2018; Rodríguez et al., 2018). Although this method has achieved good results, VR technology is not a scalable method.

Alongside these efforts to improve classical neuropsychological tests for assessing ADHD symptoms, a new generation of cognitive assessments, in the form of video games (Wiley et al., 2021), is emerging. Several serious games have been developed and validated with the aim of increasing the objectivity of the diagnosis or screening of ADHD (Penuelas-Calvo et al., 2020; Zheng et al., 2021).

In addition, the remarkable feature of serious games is that adding game elements such as graphics, storyline, scoring, feedback, and avatars can turn a long and boring assessment into an exciting and lovely experience, especially for children's age group (Khaleghi et al., 2021; Lumsden et al., 2016). Evaluation in the form of games or gamification can significantly reduce the test stress of subjects. Moreover, it can increase the subject's motivation and engagement in such a way that healthy children and children with disorders are more likely to show their true abilities during an evaluation (Lumsden et al., 2016).

Peñuelas-Calvo et. al evaluated studies on 22 video games that have been used for the assessment and treatment of ADHD in their systematic review. The results show that video games were generally useful and effective in the diagnosis and treatment of ADHD.

Unfortunately, the majority of video games designed to assess ADHD symptoms lack the assessment of hyperactivity as one of the core symptoms of this disorder.

To the best of our knowledge, among the games developed for the assessment of ADHD, only the Paediatric Attention Deficit Hyperactivity Disorder Application Software (PANDAS) game evaluates the child's motor activity during the game using tablet's accelerometer (Mwamba et al., 2019). However, in the study conducted on PANDAS, it was not determined which of the proposed motion features performed better in discriminating between healthy and ADHD children.

In this article, the validity of a serious video game, as an objective, inexpensive, and accessible method for screening children with ADHD is investigated. In addition to evaluate subjects' attention and impulsiveness during the game, it evaluates subjects' hyperactivity using accelerometers and gyroscopes of Smartphone. The game, which is called FishFinder, has been tested on a group of ADHD and healthy children with high accuracy.

**Method**

**Participants**

　　52 children between the ages of 5 and 12, in which 26 children had ADHD and 26 healthy children, participated in this study. All children were physically healthy, with no vision, hearing, movement, or any other disability, that would prevent them from playing the game.

　　Children with ADHD were selected from referrals to three child and adolescent psychiatry specialists in Tehran, Iran who had made a definite diagnosis of ADHD. Healthy children were selected through an invitation to participate in the project, which was published in physical and online forms. The healthy group participants were selected as close as possible to the ADHD group participants considering their gender and age (Table 1). Parents of all children completed an online version of the Conners' Parent Rating Scale (CPRS) and Child Behavior Checklist (CBCL). This allowed us to make sure that there was no psychological disorder in the healthy group. This phase was conducted between May 2022 and October 2022.

**Table 1**

*Details of Participants*

| Gender | | ADHD | Controls |
| --- | --- | --- | --- |
| Male | Number | 14 | 13 |
| | Age: Mean (SD) | 7.35 (2.34) | 7.07 (2.17) |
| Female | Number | 12 | 13 |
| | Age: Mean (SD) | 6.66 (2.18) | 7 (2.16) |
| Total | Number | 26 | 26 |
| | Age: Mean (SD) | 7.03 (2.25) | 7.03 (2.12) |

　　Six of the ADHD children played the game after their first visit to the psychiatrist for diagnosis before starting drug treatments. The rest of the ADHD subjects were already under drug treatment. The children under treatment were asked to stop using their treatment up to a day before the test. This process was approved by the children's psychiatrist and was conducted with the consent of their parents.

The ethical approval process was administered by the research ethics committees of the Faculty of Psychology and Education, at University of Tehran (approval ID: IR.UT.PSYEDU.REC.1401.007).

**Instrument**

***Conners' Parent Rating Scale-Revised: Short Version (CPRS-RS)***

The revised version of Conners' rating scale includes parent, teacher, and self-report forms, each with short and long versions. It is used to evaluate Attention Deficit Hyperactivity Disorder in children aged 3 to 17. The short version of Conners' Rating Scale for parents contains 27 items and is used in this study. The scale has shown good internal reliability coefficients, high test-retest reliability, and effective discriminatory power (Conners et al., 1998). The Persian version of this scale has also demonstrated good psychometric properties, proving its usefulness in distinguishing children with ADHD from typically developing individuals (Shahaeian, 2007).

***Child Behavior Checklist (CBCL)***

Child behavior checklist is a widely used questionnaire to assess children's emotional and behavioral problems. In this study, the parent rating form (PRF), consisting of 113 items, was used. Additionally, there are other versions of this form including the teacher report form (TRF), completed by teachers, and the youth self-report form (YSR), completed by the youth themselves. The questionnaire uses a three-point scale to rate items (Achenbach & Edelbrock, 1983). It has robust psychometric properties reported in various countries (Achenbach, 1991). The validation study conducted by Tehrani-Doost et al. (2011) confirmed the appropriateness of this scale to be used for the Iranian population.

**The FishFinder Game**

***The Game Structure***

Fish Finder is a video game that is designed for use on Smartphones (the most accessible digital device across the world) and developed with the Unity game engine.

This game is inspired by CPT tests. It has incorporated game design elements such as the use of a theme, storyline, animation, avatar, feedback, scoring system, time pressure, and sound effects. As a result, children are likely to play the game in a relaxed and engaging environment. Thus, it would not be stressful and boring like the typical CPT tests.

The game starts with the story of the game in the form of a one-minute animation. The game is about two siblings, who plan to dive to collect pearls, oysters, and diamonds. Suddenly they hear voices asking them for help. These are the voices of fishes that have been followed by a group of sharks. Sara and Soheil decide to help the fishes, so they chase them when they suddenly come across a strange scene. The fishes and sharks pass through a strange gate that traps them all in bubbles. Players should help the fishes by clicking on their bubbles. The main challenge they face is that they should be very careful to not release the sharks.

The overall structure of the game from this section onwards is inspired by the IVA-2 test (one of the most reliable CPT tests to measure attention and impulsivity) and has 6 sections, where the arrangement of stimuli and the ratio of target and non-target stimuli are different in these sections.

The player's task is to release fishes by bursting their bubbles but not the sharks' bubbles. Fishes and sharks are presented in both visual and auditory forms and in quasi-random order. In the auditory section, the bubbles are opaque, and it is not clear what are trapped inside them. In this situation, Sarah helps players to identify what is inside bubbles by saying shark or fish.

**Figure 1**

*FishFinder Game Graphics*

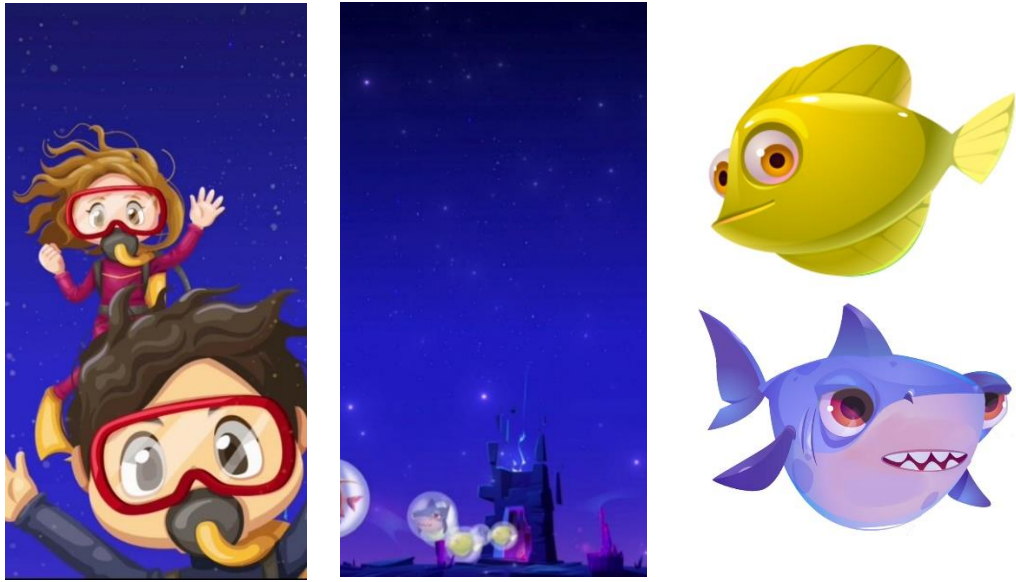

*Note.* Left panel: The characters of the game: Sarah and Soheil. middle panel: The strange gate that puts the sharks and the fishes in bubbles. right panel: The fishes and sharks in the game.

  The first section of the game is designed to familiarize players with the mechanics of the game (bubble bursting) and with the visual target stimuli (visual fish). The second section is like the first section, except that the familiarization is for the auditory part. The third section of the game is the practice section. Children receive feedback that helps them learn how to play correctly. Immediate feedback is given to users right after each mistake, in both audio form (with Sarah's voice saying: "wrong") and visual form (by changing the color of the fish icon on the top of the screen to red). Also, after three wrong selections, the game instruction menu will be displayed again for users automatically. After the practice section, the main section of the game starts. Users spend most of their time in this section (about 13 minutes), the ratio of fish to shark changes 8 times during the 50 trials to avoid learning the fish and shark appearance sequence. In this section, unlike the practice section, voice feedback is no longer given for wrong choices, and after three wrong choices, the instructions are not repeated. Like other parts of the game, players are still given visual feedback and the fish icon on the top of the screen turns red.

After completing the main part, by presenting 10 visual target stimuli (Section 5) and then 10 auditory target stimuli (Section 6), the game ends. More details about each game session are provided in Appendix A.

Before each section, the instruction of that section is shown and read to users. After completing the instructions, users are asked to tap on the star at the bottom of the menu to start the corresponding section. The player cannot enter the next section by touching the star until the end of the instructions phase. After finishing all sections, the game ends by giving three golden stars to the child and cheering with the sound of hands.

**Figure 2**

*Screenshots of the FishFinder Game*

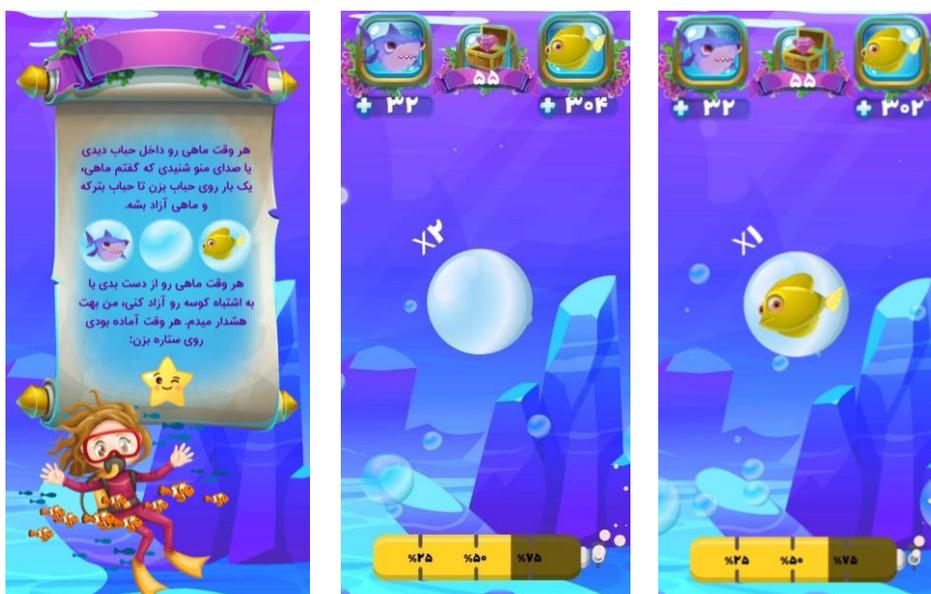

*Note.* Left panel: A view of one of the game's instruction menus. middle panel: A view of the game when presenting the auditory stimulus. right panel: A view of the game when presenting the visual stimulus.

### *Game Scoring System*

The game scoring is such that all players win. In fact, there is no option of losing the game. The game scoring is in the form of a Combo. In this way, if the child releases fishes correctly in a row and does not release a shark, he will get more points for each release of the fish. In other words, the point of the released fish multiplied by a coefficient, starting from one that can reach five at most. With the first mistake, the coefficient rewinds back to one.

If the child releases the fish at high speed (reaction time less than 400 milliseconds), in addition to the fish, a purple diamond will also come out of the bubble and be awarded to the child. Fishes, sharks, and diamonds released during the game move to their corresponding icons on the top of the screen and increase the player's score.

## Procedure

Each child and his/her parents were invited to a quiet room, where visual and auditory distractions were removed as much as possible. Then, a proper communication was established with the child through talking or giving gifts to relieve or reduce stress. The participants were told that they were invited to play a mobile game.

If the child consented, the parents were asked to leave the room while their child was playing. If the child wanted to play the game in the presence of one of the parents, this request was complied. The parent was asked to sit quietly on the chair behind the child to not distract the child. Then, the participants were invited to sit and lean on a chair without a handle in the middle of the room, and a smartphone was given to them.

Then, the participants were told that they should not talk to the examiner or their parents during the game. Also, they were assured that the game itself contained all the necessary instructions, and if they had any questions or comments, they were asked to wait and tell the examiner after finishing the game.

Then, the participants were told that whenever they were ready, they could hit the start button and start the game. The examiner sat on another chair overlooking the child and pretended that she is occupied with writing so the child would not feel under observation and be comfortable. During the game, the examiner's observations of the child's behavior were recorded.

In cases where the child left the Smartphone on the chair, the examiner immediately told the child to take the Smartphone in his/her hand and continue the game. In limited cases children got tired at the middle of the game and did not intend to continue playing. Thus, they were told that they would win by finishing the game. If a child asked the examiner a question, he/she was reminded very briefly that they should not talk to each other during the game and the child can ask his question after the game. If the child asked about the remaining time of the game, the examiner told him/her that the game will end when the air capsule under the screen runs out.

The game continued for about 20 minutes. The game's data and the data related to the movement activity were automatically recorded in a database after the game was finished.

**Figure 3**

*The Sitting Position of the Player While Playing the Game: A Healthy 5-Year-Old Boy*

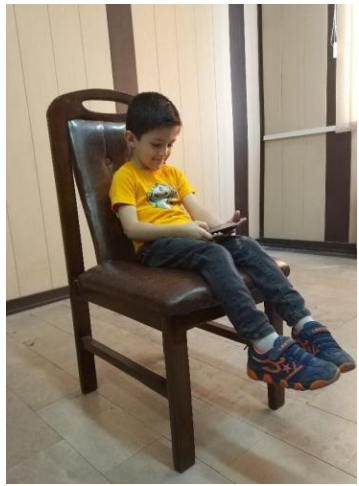

**Data Design**

Two types of data were gathered from each participant. The first dataset was the in-game data. In other words, the following data were collected during the game: Game phase (e.g., the training or the main test phase), phase mode (e.g., Fish auditory, Fish visual), starting time, reaction time to each stimulus, and the child's selections. This dataset is called the game dataset.

The Second dataset contained six signals which were recorded from the accelerometer and gyroscope of the Smartphone in each direction of x, y, and z. These signals were recorded with a frequency of 25 samples per second (Hz) and showed the movement of the Smartphone while the child was playing the game. This dataset is called the movement dataset.

**Feature Extraction**

Feature extraction is an important step in analyzing the data as it gives insight into the dataset. In addition, features are the meaning of the raw dataset which is the key information for the classifier to help it distinguish each class.

Due to type differences between these two datasets, each dataset had its own feature sets. So, there were two independent feature sets extracted from each dataset. The first one was referred to as the in-game feature set and the second one was referred to as a movement feature set. The in-game feature set was inspired by IVA-2 scale features which were 18 features extracted from auditory and visual modes (Sandford & Sandford, 2015). These features were in three categories:

(a) Response Control features including Prudence, Consistency, and Stamina

(b) Attention features including Vigilance, Focus, and Speed

(c) Symptomatic features including Comprehension, Persistence, and Sensory

The movement features were extracted after a preprocessing phase in which the first 15 seconds and the last 60 seconds of the movement data for each subject were deleted. This is done due to the fact that in these two time-intervals, the Smartphone movement was due to the interaction between the examiner and subjects. Furthermore, we only used the movement data of the second half of the game since we observed that ADHD children tend to show more movement in the second part compared to the first one. The movement feature set contains 96 features which can be divided into three categories. They are described in Table 2.

**Table 2**

*The Movement Features*

| Feature category | Explanation |
|---|---|
| Time domain features | Statistical features including variance, mean, maximum, minimum, range, median, summation, skewness that were calculated only for the accelerometer in x, y, and z axis and the amplitude, mean of absolute values and standard deviation of absolute values that were calculated only for gyroscope in x, y, and z axis, and standard deviation that was calculated for both accelerometer and gyroscope in x, y, and z axis and amplitude. |
| Frequency domain features | These features were extracted for only the accelerometer for each x, y, and z axis: first five frequencies, first five FFT amplitudes, energy of the signal, and pair correlation between each two axes. |
| Our proposed features | These features are aimed to show the oscillation of the signal and includes the features that were the innovation of this study. They are described in Table 3. |

Our proposed features are described in Table 3.

**Table 3**

*Our Proposed Features*

| Feature | Explanation | Number of Features |
|---|---|---|
| Active (x, y, z) | Percentage of values above the specified threshold | 3 |
| Inactive (x, y, z) | Percentage of values under the specified threshold | 3 |
| Big (x, y, z, amplitude) | The number of oscillations with an amplitude greater than the specified value | 4 |
| Active Max | Maximum of Active values | 1 |
| Active Mean | Mean of Active values | 1 |
| Big Max | Maximum of Big values | 1 |
| Big Mean | Mean of Big values | 1 |

After extracting all the features, they were scaled using the MinMaxScaler function in the Sklearn library which changes the scale of each feature to the range of 0 to 1 without changing its distribution. The Performance of this scaler is to subtract the minimum value in a feature and then divide it by the range. Scaling feature is a necessary modification to improve the efficiency of the classifier.

**Support Vector Machine**

Recognizing distinctive movement or game patterns between ADHD and normal groups is possible after extracting the feature vector from each sample. The samples are divided into train and test groups. The training group, consisting of 50% of all the samples is used to train the classifier. In the next step the data from the test group which is the remaining 50% of the samples, is used to measure the accuracy of the designed classifier.

A Support Vector Machine (SVM) classifier, with a soft margin and linear kernel, is used for classification purpose.

**Feature Selection**

Due to the small sample size and the large number of features, selecting the most relevant features for training a classifier is important to leave out unnecessary and additional information. Three feature selection methods were tried:

a) correlation: eliminating one of the two correlated features

b) forward selection: to find the most important features

c) backward elimination: to remove the least important features.

Each of these methods was applied to the in-game feature set and movement feature set independently.

At the end and for selecting the final feature set, the features that were selected in more than two methods were chosen to train the classifier.

The results of the classifier on each feature selection method using the in-game feature set are determined in Table 4.

**Table 4**

*The Result of the Classifiers With Different Feature Conditioning Methods on Game Feature*

| Methods | Test Accuracy | Sensitivity | Specificity | Number of Features |
|---|---|---|---|---|
| All | 0.81 | 0.90 | 0.75 | 18 |
| Correlation | 0.81 | 0.90 | 0.75 | 2 |
| Forward Selection | 0.88 | 1 | 0.81 | 7 |
| Backward Elimination | 0.85 | 0.90 | 0.81 | 4 |

A similar feature selection was applied to movement features. Furthermore, feature elimination was used to select the most effective movement features, and the results are presented in Table 5.

**Table 5**

*Final Movement Features Results*

| Test Accuracy | Sensitivity | Specificity | Number of Features |
|---|---|---|---|
| 0.88 | 0.80 | 0.94 | 5 |

## Results

Considering the final selected features of each feature set, in the final stage, the classifier was trained on both the final selected in-game and movement features.

The final features from the in-game feature set are:

- Auditory and visual Consistency: This feature evaluates the general reliability and variability of response times. It measures the ability to stay on task.

- Auditory and visual stamina: This feature compares the average reaction times of accurate responses in the initial 200 trials with those in the last 200 trials and helps to detect issues related to sustained attention and effort over time.

- Auditory vigilance: This feature evaluates inattention by analyzing two distinct types of omission errors,

- Visual comprehension: This feature detects random responding by measuring idiopathic errors. Research has shown that this sub-scale is the most sensitive in effectively distinguishing ADHD (Sandford & Sandford, 2015).

The final features from the movement feature set are: The Standard deviation of the gyroscope in the x-axis, the Active feature of the gyroscope in the x-axis, the range of the accelerometer in the x-axis, the first FFT amplitude of the accelerometer in the x-axis, and the variance of the accelerometer in the x-axis.

The result of classification on both the final in-game and movement features is shown in Table 6.

**Table 6**

*Performance metrics of SVM when the in-game and the movement features are combined*

| Metric | Equation | Value |
|---|---|---|
| Test Accuracy | (TP+TN)/(TP+TN+FP+FN) | 0.92 |
| Sensitivity | TP/(TP+FN) | 0.90 |
| Specificity | TN/(TN+FP) | 0.94 |
| Precision | TP/(TP+FP) | 0.90 |
| F1Score | (2×TP)/(TP+FP+FN) | 0.90 |

**Figure 4**

*Confusion Matrix of Final Classifier*

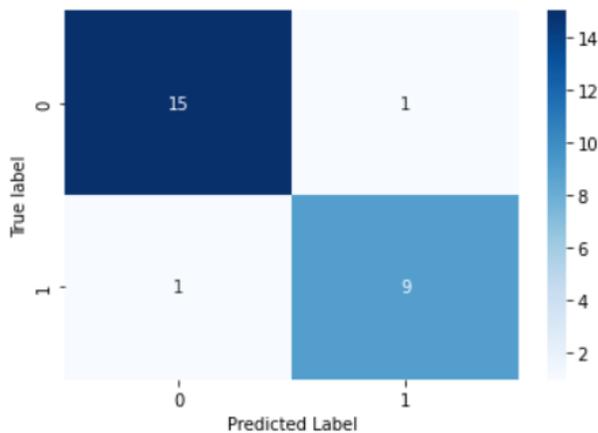

Finally, to facilitate the comparison of three different modes of the model, the results of mode 1 (i.e., by using only the in-game features), mode 2 (i.e., by using only the movement features), and mode 3 (i.e., by using all features), are presented in Table 7.

**Table 7**

*Performance Metrics of Support Vector Machine (SVM) In Three Modes*

| Metric | Mode1 | Mode2 | Mode 3 |
| --- | --- | --- | --- |
| Test Accuracy | 0.96 | 0.89 | 0.92 |
| Sensitivity | 0.90 | 0.80 | 0.90 |
| Specificity | 1 | 0.94 | 0.94 |
| Precision | 1 | 0.88 | 0.90 |
| F1Score | 1 | 0.84 | 0.90 |
| Number of Features | 6 | 5 | 11 |

## Discussion

In this study, we evaluated the validity of the FishFinder game in children's ADHD screening. The accuracy of the game while using two different sources of information which are the in-game data and the movement features, are reported. The accuracy of screening while using only the in-game features (mode 1) was 96.1%, using only the motion features (mode 2) was 88.4%, and using all features together (mode 3) was 92.3%. It can be seen that screening children using the in-game features and all features together can be done with very good accuracy.

To explain the difference between the lower accuracy of mode 2 with the other two modes, it should be said that the amount of movement and vitality of ADHD (and healthy) children is greatly influenced by their age (Andrés Martin et al., 2018). However, in the current study, the number of samples is not sufficient to allow us to have different groups of various ages. Thus, children of different ages are not separated in this study. Although the results are satisfactory when these children are put in one group, it is expected that by increasing the sample size including enough samples of girls and boys of different ages, the accuracy of the model, its sensitivity, and specificity be higher than the current values. As it is predicted that the rate of false positives and false negatives decreases by doing so. Currently, it is possible that healthy children at lower ages fall into the false positive category due to their higher motor activity which is normal. Moreover, children with ADHD at older ages fall into the false negative category due to the reduction of their hyperactivity symptoms over time.

In the following, Fish Finder is compared with other serious games or common neuropsychological tests to distinguish ADHD children from healthy children.

**Comparison With Similar Serious Games**

Consistent with previous literature on serious games for screening and assessing ADHD symptoms (Penuelas-Calvo et al., 2020), the results show that a video game can successfully measure ADHD symptoms. The advantage of video games compared to traditional screening/ assessment tools is in the more enjoyment while playing and reducing their stress. The high accuracy of the FishFinder game in children's ADHD screening is very satisfactory compared to other games. The sensitivity and specificity of FishFinder game are also greater than or equal to the existing serious games. In Table 8 the summary of the sensitivity and specificity of several successful serious games developed to measure ADHD symptoms are listed alongside FishFinder game.

**Table 8**

*Comparison of the FishFinder Game Classification Performance With Existing Serious Games*

|  | FishFinder | Groundskeeper | Timo's Adventure | Name not specified | PANDAS | PANDAS |
|---|---|---|---|---|---|---|
| Study | Current study | Heller et al. (2013) | Peijnenborgh et al. (2016) | Pires et al. (2015) | Mwamba et al. (2019) | Swarts et al. (2019) |
| Sensitivity | %90 | %76.9 | %89 | %82.4 | %75 | %90.7 |
| Specificity | %93.7 | %80.7 | %69 | %82.4 | %100 | %94.4 |
| Number of ADHD vs. Total | 52 (26) | 52 (26) | 97 (40) | 34 (17) | 30 (11) | 39 (8) |

The better results of the FishFinder game can be due to its design based on valid CPT tests which can make it possible to clearly observe ADHD symptoms.

**Comparison With Common CPT Tests**

Fish Finder is a cognitive game and not a simple cognitive test. Considering that it has been inspired from continuous performance tests, there are many similarities between the features extracted from the game and the output indicators of CPT tests. Thus, its results can be compared to common CPT tests.

Common CPT tests do not measure motor activity. Consequently, the movement features of the FishFinder game cannot be compared to these tests. In other words, only the performance of the FishFinder game in mode 1 can be compared to these tests. The high sensitivity and specificity of the FishFinder game based on features related to attention and impulsivity in ADHD screening are very satisfactory and equal or superior to the existing continuous performance tests. Table 9 summarizes the sensitivity and specificity of some of the most important CPT tests that have been employed to measure attention and impulsivity in ADHD.

**Table 9**

*Comparison of the FishFinder Game Classification Performance With CPT Tests*

|  | FishFinder | IVA | TOVA | CCPT |
|---|---|---|---|---|
| Study | Current study | Sandford & Sandford (2015) | Schatz et al. (2001) | Perugini et al. (2000) |
| Sensitivity | %90 | %92 | %85.7 | %67 |
| Specificity | %100 | %90 | %70 | %73 |
| Number of ADHD vs. Total | 52 (26) | 57 (26) | 48 (28) | 43 (21) (Boy) |

The better performance of the Fish Finder game compared to the well-known CPT tests can be due to the following reasons. Firstly, the use of various distracting stimuli throughout the game (Animated bubbles, change in the child's scores, and reduction of the oxygen capsule), leads to more challenges for ADHD children and thus helps to better distinguish them from healthy children. Secondly, the auditory attention and response control along with visual attention and response control are assessed simultaneously (The auditory mode is not present in common CPT tests but is seen in the IVA test).

**Comparison With CPT Tests With Simultaneous Measurement of Motor Activity**

As mentioned before, neither the available serious games (except the PANDAS game) nor the majority of CPT tests, do not measure hyperactivity. However, the Fish-Finder game benefits from the assessment of child movement as well.

The use of motion features alone showed a satisfactory ability of the FishFinder game to distinguish these children from healthy children (Accuracy %88.4, Sensitivity %80, Specificity %93.7). This is because these children, due to their hyperactivity and restlessness have more movement activity during the game. For example, they move more when seated, move their legs back and forth, move their Smartphone in their hands, and play with their hair or clothes. Also, in cases with more severe symptoms of hyperactivity, occasionally, they suddenly stand up and then return to their place again. Table 10 shows the summary of the sensitivity and specificity of some of the most important CPT tests that have been developed to measure hyperactivity along with attention and impulsivity.

The sensitivity and specificity of the FishFinder game in the screening of children with ADHD is equal to or higher than the existing methods of simultaneous objective measurement of the three core symptoms of this disorder.

The important advantage of the FishFinder game compared to these tests is the use of motion sensors in Smartphones instead of using fewer common technologies such as using infrared camera, virtual reality, and webcam.

Consequently, the FishFinder game is an inexpensive objective screening method that can be accessible on a large scale.

**Table 10**

*Comparison of the FishFinder Game Classification Performance With CPT Tests That Assess*

*Motor Activity Simultaneously*

|  | FishFinder | Aula | MMAT | QbTest | Qbcheck |
|---|---|---|---|---|---|
| Study | Current study | Rufo-Campos et al. (2012), as cited in Kane & Parsons, (2017) | TEICHER et al. (1996) | Emser et al. (2018) | Ulberstad et al. (2020) |
| Sensitivity | %90 | %95.2 | %88.9 | %80 | %82.6 |
| Specificity | %93.7 | %91.8 | %100 | %77 | %79.5 |
| Number of ADHD vs. Total | 52 (26) | 124 (62) | 29 (18) (Boy) | 60 (30) | 142 (69) (Adult) |
| Hyperactivity assessment technology | Smartphone accelerometer and gyroscope | Movement sensors placed in the virtual reality headset | Infrared camera | Infrared camera | Webcam |

This study has several limitations. The use of drug-medicated children in the group of people with ADHD should be considered as one of the limitations. To decrease the short-term effects of using drugs, these children took the dose of their medicine after playing the game instead of having it before (this action was done in coordination with their child and adolescent psychiatrist). However, the long-term effects of the medication were still present in these children, and they showed less severe symptoms compared to the ADHD children who had not taken any medication. It is suggested that for future research on the FishFinder game, ADHD samples that have not been treated so far be selected to the greatest extent.

Another limitation of this study was not having separate samples of different ages and genders. This issue also greatly impacts the ability of movement features to distinguish between healthy children and ADHD. Since the amount of movement activity of children at different ages and between girls and boys varies obviously. For example, a healthy 5-year-old boy may have more vitality than a 12-year-old boy with ADHD, and this explains why children need to be compared to their age and gender groups. It should be noted that the features related to attention and impulsivity showed a very high ability to separate healthy and ADHD children without considering separate samples for different genders and ages.

Along with the limitations, the results show several strengths of the FishFinder game. Firstly, evaluating all three core symptoms of ADHD (i.e., inattention, hyperactivity, and impulsivity) simultaneously has empowered the prediction results. Simultaneous measurement of hyperactivity is a valuable feature that distinguishes this game from similar serious games and even many existing neuropsychological tests that are used to evaluate the symptoms of ADHD. Also, two motion features used in this game (the active feature of the gyroscope in the x-axis and the first five frequencies of the accelerometer in the x-axis) are proposed for the first time in this study.

the FishFinder game can be used as an objective screening method for ADHD without the need for subjective reports from parents and teachers. Considering the heavy reliance of the current common assessment and diagnosis of ADHD on subjective reports and the errors that may occur in this process, using this game eliminates these kinds of errors.

Another important strength of the proposed method is taking advantage of using game elements. FishFnder is a serious game that screens children without they realize that they are being evaluated. Consequently, it lowers children's stress and increases his motivation to continue and finish the task. The result of all these cases compared to a simple and usually boring cognitive test is that children (both healthy and ADHD children) are expected to show their highest performance.

Finally, it should be said that the FishFinder game can be used as an accessible, inexpensive, and scalable method for objective screening of ADHD because it is designed to be used on a Smartphone (the most accessible digital device) and the game can be employed without the presence of a specialist.

**Conclusions**

In this article, the use of a serious video game along with the assessment of children's movement activity during the game using smartphone accelerometer and gyroscope sensors to screen ADHD in children aged 5 to 12 years is discussed.

In the case of combining in-game features and movement features (i.e., inattention, hyperactivity, and impulsivity which are three core symptoms of the disorder), the accuracy of 92.3%, sensitivity of 90%, and specificity of 93.7% were obtained, which is very satisfactory and shows the high ability of the FishFinder game in differentiating ADHD children from healthy children. The FishFinder game is an affordable, accessible, and attractive screening method that can be widely used on a large scale.

**Appendix A**
**Details of FishFinder Game Sections**

| number | section | Description |
|:---:|:---:|:---|
| 1 | Game story animation | When the child presses the button to start the game, in the first step, the story of the game is played for the child in the form of a one-minute animation. |
| 2 | Initial (visual) | 10 target stimuli (fish only) are presented in visual mode |
| 3 | Initial (auditory) | 10 target stimuli (fish only) are presented in auditory mode |
| 4 | Practice | It contains 32 trials, which are a combination of the target (fish) and non-target (shark) stimuli, which are presented in two visual and auditory modes. |
| | | By making three mistakes (releasing the shark or not releasing the fish), the child receives feedback, and the instructions for the game are shown again to ensure that the child is familiar with the correct way to play the game before starting the main part. |
| 5 | main | This section contains eight parts, and each part contains 50 trials. At this stage, a total of 400 stimuli are presented. |
| | | In parts one, three, five, and seven of this section, there are 42 target stimuli and eight non-target stimuli. |
| | | In parts two, four, six, and eight, this ratio is reversed and there are eight target stimuli and 42 non-target stimuli. |
| 6 | Final (visual) | 10 target stimuli (fish only) are presented in visual mode. |
| 7 | Final (auditory) | 10 target stimuli (fish only) are presented in auditory mode |
| 8 | Encourage the player | At the end of the game, Sarah's character thanks the child for the player's help in saving the fish, and when the player is announced as a winner, the child receives three golden stars and is applauded. |